%% file: main.tex
\documentclass[twocolumn,twoside]{IEEEtran}
\usepackage{graphicx}

\ifCLASSINFOpdf

\else

\fi

\usepackage[dvipsnames]{xcolor}
\usepackage{threeparttable}
\usepackage{tabularx} % for adjustable column width
\usepackage{booktabs}
\usepackage{stfloats}
\usepackage{flushend}
\definecolor{darkgreen}{rgb}{0,0.6,0}
\usepackage[bookmarks=true,colorlinks=true,pdfpagemode=UseNone,citecolor=darkgreen,linkcolor=blue,urlcolor=blue]{hyperref}

\begin{document}
\bstctlcite{MyControl}

\title{A Survey of Legged Robotics in Non-Inertial Environments: Past, Present, and Future\thanks{\em Corresponding authors: Jingang Yi and Yan Gu.}}

\author{
I-Chia Chang\thanks{I.-C. Chang, W. Li, and Y. Gu are with the School of Mechanical Engineering, Purdue University, West Lafayette, IN 47906 USA (e-mail: \{chang970,li5923,yangu\}@purdue.edu).}, Xinyan Huang\thanks{X. Huang and J. Yi are with the Department of Mechanical and Aerospace Engineering, Rutgers University, Piscataway, NJ 08854 USA (e-mail: \{xh301,jgyi\}@rutgers.edu).}, Tzu-Yuan Lin\thanks{T.-Y. Lin, S. Teng, and M. Ghaffari are with the Department of Robotics, University of Michigan, Ann Arbor, MI 48109 USA (e-mail: \{tzuyuan,sanglit,maanigj\}@umich.edu).}, Sangli Teng, Wenjing Li, Maani Ghaffari, Jingang Yi, Yan Gu
}

\maketitle

\begin{abstract}

Legged robots have demonstrated remarkable agility on rigid, stationary ground, but their locomotion reliability remains limited in non-inertial environments, where the supporting ground moves, tilts, or accelerates. Such conditions arise in ground transportation, maritime platforms, and aerospace settings, and they introduce persistent time-varying disturbances that break the stationary-ground assumptions underlying conventional legged locomotion. This survey reviews the state of the art in modeling, state estimation, and control for legged robots in non-inertial environments. We summarize representative application domains and motion characteristics, analyze the root causes of locomotion performance degradation, and review existing methods together with their key assumptions and limitations. We further identify open problems in robot-environment coupling, observability, robustness, and experimental validation, and discuss future directions in autonomy, system-level design, bio-inspired strategies, safety, and testing. The survey aims to clarify the technical foundations of this emerging area and support the development of reliable legged robots for real-world dynamic environments.

\end{abstract}


\section{Introduction}

Legged robots offer mobility on terrains that wheeled and tracked systems cannot easily traverse. However, their reliability drops sharply in \emph{non-inertial} environments, where the supporting surface accelerates relative to the inertial Earth frame, such as on moving ships, trains, or aircraft. In these settings, time-varying platform motion continuously perturbs robot balance, contact, and perception, making robust locomotion substantially more challenging than on stationary ground.

\subsection{Why Do Non-Inertial Environments Matter?}

Non-inertial environments are common in transportation and industrial settings across ground, maritime, and aerospace domains.
In the ground domain, public transportation systems such as buses, subways, and trains carried more than 7.1 billion trips in the United States in 2023~\cite{APTA_2024}, motivating robotic support for passenger safety, surveillance, and cleaning in confined vehicles. In the maritime domain, ships and offshore platforms account for more than 80\% of global trade by volume~\cite{shipping-data-unctad-2025} and serve more than 30 million cruise passengers annually \cite{2024_Cruise_Line}, creating strong demand for robotic inspection, maintenance, and emergency response under harsh sea states. In aerospace, commercial airlines transported 5 billion passengers worldwide in 2024 \cite{IATA_Safety}, while turbulence-related incidents continue to cause serious injuries \cite{Katz_Emont_Solomon_2024}, highlighting the potential value of legged robots for in-flight service, inspection, cargo stabilization, and post-incident assistance.

\begin{figure}
    \centering
    \includegraphics[width=1\linewidth]{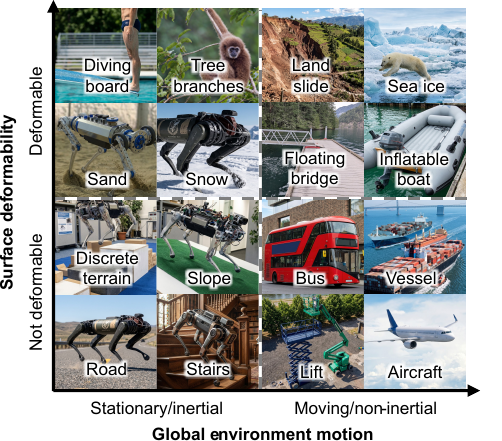}
    \vspace{-0.15 in}
    \caption{Category of different terrains and environments
    spanned by the axis of surface deformability, from rigid to deformable surfaces, and the axis of environment motion, from inertial to non-inertial environments.}
    \label{fig: category of environment}
\end{figure}

\subsection{Research Challenges and Gaps}

Non-inertial environments fundamentally break the stationary-ground assumptions underlying conventional legged robot modeling, state estimation, and control. Relative to stationary ground, reliable locomotion in such environments is more challenging because platform accelerations introduce unknown, time-varying, and often multidirectional disturbances. Recent studies~\cite{misenti_experimental_2025,gao_time-varying_2024} have revealed clear performance gaps in legged robot systems operating on moving platforms. For example, the proprietary systems of commercial robots such as Boston Dynamics' Spot exhibit significant body-position drift even when stepping in place on a wave-disturbed ship under mild sea conditions. 
Unlike impulsive disturbances such as external pushes, non-inertial effects are persistent and continuously challenge balance. Therefore, achieving reliable legged locomotion in non-inertial environments remains a fundamental challenge in robot modeling, state estimation, and control.

\begin{figure*}[t]
    \centering
    \includegraphics[width=0.95\linewidth]{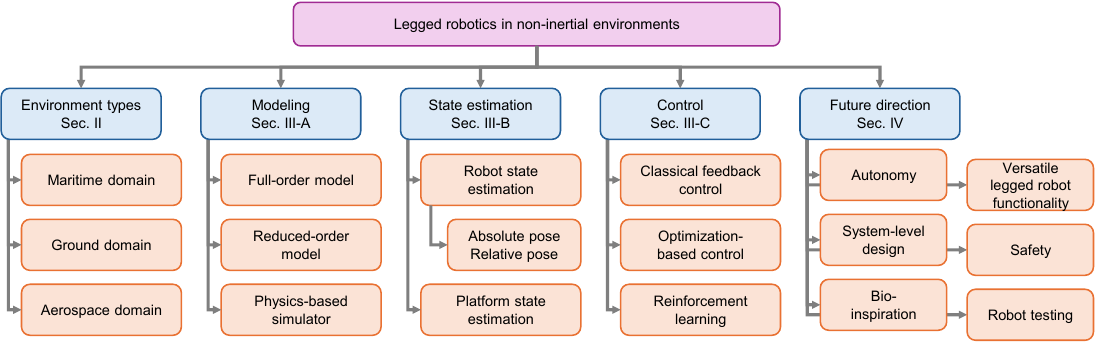}
    \vspace{-0.1 in}
    \caption{Paper overview and taxonomy of legged robotics in non-inertial environments.}
    \label{fig:placeholder}
\end{figure*}

\subsection{Scope and Focus of the Survey}

This survey reviews the existing research on modeling, state estimation, and control, including both classical and learning-based methods, for legged locomotion in non-inertial environments. We also discuss future directions in autonomy, system design, and safety. We focus on modeling, state estimation, and control because they form the algorithmic foundation of legged locomotion and directly shape stability, robustness, agility, and versatility. Motion planning, such as footstep generation, is discussed within the broader context of control because it directly regulates locomotion behavior. Beyond the technical review, this survey also highlights the practical importance of robotic applications in non-inertial environments and identifies open challenges and future research directions.

To avoid ambiguity, we clarify the scope of the term ``non-inertial'' in this survey. Here, it does not refer to the relativistic interpretation in general relativity, zero- or microgravity conditions, or underwater settings. We further distinguish between body-attached non-inertial frames, such as a frame attached to a robot's torso, and ground-attached non-inertial frames, with the latter being the focus of this survey. While body-attached non-inertiality arises from relative motion among body links, ground-attached non-inertiality arises from motion of the supporting surface relative to the inertial Earth frame. This invalidates standard stationary-ground assumptions in modeling, estimation, and control. We also note that non-inertiality and surface deformability are distinct environmental properties, as illustrated in Fig.~\ref{fig: category of environment}. For example, a floating boat with a soft deck may be both deformable and non-inertial.

Existing surveys on legged locomotion primarily focus on control methods for stationary, inertial environments \cite{reher2021dynamic, wensing2023optimization, ha2024learning, gu2025evolution}. By contrast, this paper examines the distinct algorithmic and mechatronic challenges introduced by non-inertial environments, including fictitious forces and base-acceleration coupling. More broadly, although legged robots in unstructured or extreme environments have attracted substantial attention~\cite{wong2018autonomous}, prior reviews have largely treated ``challenging terrain'' as stationary irregular surfaces~\cite{torres2022legged}, non-Newtonian fluids~\cite{godon2023maneuvering}, or specialized domains such as subsea~\cite{picardi2023underwater}, space~\cite{gao2017review}, and agricultural environments~\cite{etezadi2024comprehensive}. The dynamic complexities of locomotion on accelerating substrates remain under-represented, which motivates this survey.

\subsection{Contributions}
This survey makes the following contributions:
\begin{enumerate}
\item [(a)] To the best of our knowledge, it is the first survey focused on legged robots, and robotics more broadly, in non-inertial environments.
\item [(b)] It provides a first-principles perspective on the fundamental challenges that non-inertial environments introduce to legged locomotion, clarifying the root causes of current failures while guiding solution development.
\item [(c)] It reviews the state of the art in modeling, state estimation, and control, and highlights the main limitations and open gaps in existing approaches.
\item [(d)] It synthesizes open challenges and future directions in autonomy, system-level design, bio-inspiration, loco-manipulation, human-robot interaction, safety, and robot testing, providing a coherent road map for advancing robotics in non-inertial settings.
\end{enumerate}

The remainder of this paper is organized as follows (see Fig.~\ref{fig:placeholder}). 
Section~\ref{sec:domainchallenges} reviews the motion characteristics and application-driven challenges of representative non-inertial domains. 
Section~\ref{sec:technical foundations} reviews existing approaches in modeling, state estimation, and control, and identifies associated open problems. Section~\ref{sec:research challenges and future} discusses emerging future directions and Section~\ref{sec:conclusion} concludes the paper.

\section{Real-World and Application-Driven Challenges}
\label{sec:domainchallenges}

Driven by rapid advances in control, optimization, learning, and hardware, legged robots are moving steadily from laboratory demonstrations toward real-world deployment. However, most existing demonstrations have been conducted on stationary surfaces, and stable, high-performance locomotion in non-inertial environments remains largely underexplored.

\begin{figure}[t!]
    \centering
    \includegraphics[width=1\linewidth]{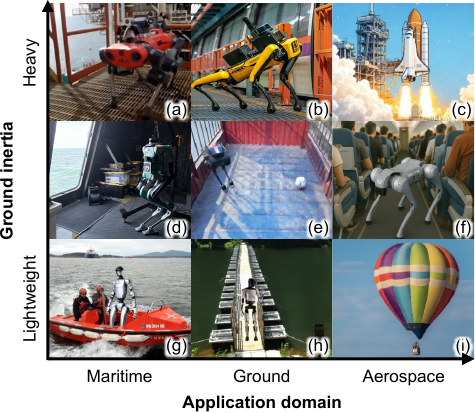}
    \vspace{-0.2 in}
    \caption{Different types of non-inertial platforms across application domains. (a) Offshore oil-platform \cite{yt_demo}. (b) Subway. (c) Spacecraft. (d) Ship cabin. (e) Moving truck \cite{bili_demo}.  (f) Airplane cabin. (g) Lightweight rescue boat. (h) Passively oscillating floating bridge. (i) Hot-air balloon. These scenarios span a broad range of non-inertial platforms, from lightweight to heavy, that are relevant to legged robot operation in real-world dynamic environments.}
    \label{fig: application domain}
\end{figure}

Such environments are nevertheless common, arising in ground transportation, maritime platforms, and aerial domains (see Fig.~\ref{fig: application domain}). From the perspective of robot-environment coupling, these platforms can be broadly classified as either heavy or lightweight. For heavy platforms, the platform motion is treated as exogenous and unaffected by the robot dynamics. For lightweight platforms, by contrast, the platform trajectory can be significantly influenced by the interaction forces generated by the robot. Beyond this distinction, non-inertial platforms also differ substantially in the magnitude, frequency, and pattern of motion. These differences give rise to distinct challenges for robust legged locomotion, as discussed next.

\subsection{Ground Domain}

Ground vehicles are the most common non-inertial environments and include buses, trains, trucks, and mobile cranes. As summarized in Table~\ref{tab:domain_motion_characteristics}, their motion combines vibration-induced body accelerations with maneuver-induced longitudinal accelerations. Overall, they exhibit vertical moderate-frequency vibrations with potentially significant transient horizontal accelerations. Under normal urban and highway driving (50--110\,km/h), passenger vehicles exhibit dominant vibrations in the 1--5\,Hz range, with wheel-road interaction bands extending to 5--25\,Hz. The resulting body acceleration amplitudes are typically up to 0.12\,m/s$^2$ on well-maintained roads and up to 0.18\,m/s$^2$ on rougher roads, together with low-amplitude roll and pitch~\cite{praznowski_assessment_2020}. In railway vehicles, car-body vertical acceleration is dominated by frequencies below 10\,Hz, with specific peaks depending on track and vehicle conditions~\cite{tsunashima-2020}. Maneuver-induced longitudinal accelerations are often larger than road-induced vibration: for passenger cars, acceleration can reach 2.9\,m/s$^2$ and deceleration 4.5\,m/s$^2$, while trucks, motorized two-wheelers, and three-wheelers show comparable order-of-magnitude values depending on vehicle type and operating condition~\cite{bokare2017acceleration}.

For legged robots, these motions create two main challenges. First, sustained vertical vibrations in the 1--5\,Hz range continuously perturb center-of-mass dynamics and contact forces, requiring robust balance control. Second, maneuver-induced longitudinal accelerations generate substantial inertial forces that shift the effective support region and demand rapid adaptation of foot placement and posture.
These two effects call for disturbance-aware planning and control to ensure reliable locomotion on ground vehicles.

\begin{table*}
	\centering
	\caption{Representative Motion Characteristics of Non-Inertial Platforms in Maritime, Ground, and Aerospace Domains.}
	\label{tab:domain_motion_characteristics}
	\renewcommand{\arraystretch}{1.25}
	\begin{tabular}{c p{4.2cm} p{5.0cm} p{6cm}}
		\hline
		\textbf{Domain} & \textbf{Platform type} & \textbf{Representative operating conditions} & \textbf{Representative motion characteristics} 
        % & \textbf{Citations }
        \\
		\hline

	{\textbf{Ground}} 
		&Passenger car
		&
		Urban and highway roads; speeds 50-110\,km/h;
		Dominant vibration content at 1--5\,Hz; wheel-road interaction bands at 5--25\,Hz
		&
		Typical vibration-induced body acceleration amplitudes:
		up to $0.12$ m/s$^2$ (well-maintained roads),
		up to $0.18$ m/s$^2$ (rougher roads);
		low-amplitude roll/pitch variations~\cite{praznowski_assessment_2020}
		\\
        \cline{3-4}
		&  
		& Urban/highway roads, normal driving maneuvers
		& Acceleration up to 2.9\,m/s$^2$; deceleration up to 4.5 m/s$^2$~\cite{bokare2017acceleration}\\
		\cline{2-4}
		
		& Truck
		& Free-flow road, normal maneuvers
		&  Acceleration up to 1.0\,m/s$^2$; deceleration up to 0.9 m/s$^2$~\cite{bokare2017acceleration}\\
		\cline{2-4}
		& Motorized three-wheeler
		& Urban/free-flow roads, normal maneuvers
		& Acceleration up to 0.6\,m/s$^2$; deceleration up to 1.1\,m/s$^2$~\cite{bokare2017acceleration}\\
		\cline{2-4}
		& Motorized two-wheeler 
		& Urban/free-flow roads, normal maneuvers
		& Acceleration up to 2.0\,m/s$^2$; deceleration up to 1.6\,m/s$^2$~\cite{bokare2017acceleration}\\

		\hline

		{\textbf{Maritime}} 
		& Trimaran semi-submersible
		& Multiple heading directions relative to waves
		& Heave $\approx 3~\mathrm{m/m}$ at $0.35~\mathrm{Hz}$;
        surge/sway $\approx 8~\mathrm{m/m}$;
        pitch $\approx 40^\circ/\mathrm{m}$ at $0.56$--$0.64~\mathrm{Hz}$;
        roll $\approx 9^\circ/\mathrm{m}$; yaw $\approx 13^\circ/\mathrm{m}$
        (RAOs, response amplitude per unit wave height; $\mathrm{m/m}$ for displacement, $^\circ/\mathrm{m}$ for rotation)~\cite{putrananda24numerical}
		\\
		\cline{2-4}
		& Large unmanned surface vessel
		&
        Sea States 5--7
		&
        Wave period $\approx 9.6$\,s;
        % ($\approx 0.65$ rad/s);  
		wave height $\approx 9$\,m.
		Heave $\pm 5$\,m;  
		pitch $\approx \pm 12^{\circ}$;  
		roll from $-7^{\circ}$ to $6^{\circ}$ depending on heading~\cite{huang2021numerical}
		\\
		\cline{2-4}
		
		& Water ambulance ship
		&
		Operating conditions under different ship loading configurations
		&
        Heave peak at $0.037$--$0.043$\,Hz;
		roll peak at $0.08$--$0.09$\,Hz.
		Typical peak amplitudes:  
		heave $\approx 0.25$--$0.28$ m;  
		pitch $\approx 0.4$--$0.9^{\circ}$;  
		roll RMS $\approx 1.6$--$1.7^{\circ}$~\cite{Alamsyah2023Motion}
        \\

		\hline
		
		{\textbf{Aerospace}} 
		& Commercial fixed-wing aircraft
		& Climb and descent 
		&  Typical linear acceleration ranges:
        vertical $0.01$--$0.03$\,m/s$^2$;
        longitudinal $0.1$--$0.3$\,m/s$^2$~\cite{elmaghraby2020normal}\\
		\cline{2-4}
		& Fixed-wing aircraft (Airbus A319)
        &
        Takeoff
        &
        Pitch angle at lift-off:
        $\approx 4.48^\circ$--$12.4^\circ$
        (optimal $\approx 9.76^\circ$)~\cite{zhang2025research}\\

        \cline{2-4}
        &Helicopter (UH-60 class)
        &
        Hover near mountainous terrain; 20-knot wind;
        terrain-induced turbulent shear layer
        &
        Pitch angle $\approx 0^\circ$--$7^\circ$;
        roll angle $\approx -15^\circ$ to $4^\circ$
        during 30\,s of stabilized hovering under strong turbulence~\cite{watson2025potential}\\
		
		\hline
	\end{tabular}
\end{table*}

\subsection{Maritime Domain}

Maritime platforms such as ships and offshore structures undergo wave-driven six-degree-of-freedom (6-DoF) motions, including roll, pitch, yaw, heave, surge, and sway. As summarized in Table~\ref{tab:domain_motion_characteristics}, these motions are typically low-frequency but large-amplitude. For example, trimaran semi-submersibles exhibit heave resonance around $0.35$\,Hz and pitch peaks near $0.56$--$0.64$\,Hz, with surge and sway responses up to $8$\,m
% m/m
and yaw up to $13^\circ$ in response amplitude operators (RAOs)~\cite{putrananda24numerical}.
Large unmanned surface vessels in Sea States 5--7 encounter wave periods around $9.6$\,s 
% (approximately $0.65$~rad/s),
(approximately $0.10$\,Hz),
with heave amplitudes up to $\pm 5$\,m, pitch angles around $\pm 12^\circ$, and roll ranging from $-7^\circ$ to $6^\circ$ depending on heading~\cite{huang2021numerical}.
Smaller vessels, such as water ambulance ships, show roll peaks at
% $0.47$--$0.56$~rad/s 
$0.08$--$0.09$\,Hz
and typical heave amplitudes of $0.25$--$0.28$\,m, with roll root-mean-square (RMS) values around $1.6$--$1.7^\circ$ \cite{Alamsyah2023Motion}. 
In summary, maritime motion is dominated by low-frequency oscillations (approximately $0.05$--$0.6$\,Hz) with substantial translational and rotational excursions.

For legged robots, these motions impose persistent, large-scale disturbances. Ship decks can undergo low-frequency roll and pitch oscillations of several to double-digit degrees, together with meter-scale heave and sway. These disturbances continuously alter the gravity direction felt by the robot, shift the effective support region, and generate significant inertial loads. Long wave periods, such as $8$--$10$\,s, preclude quasi-stationary compensation and require predictive adaptation to cyclic deck motion. On small vessels, robot locomotion can further perturb the platform, creating strong bidirectional coupling. Reliable maritime legged robot deployment thus requires disturbance-aware planning, attitude-compensated whole-body control, and, in some cases, coupled robot-platform modeling.

\subsection{Aerospace Domain}

Aircraft platforms exhibit distinct non-inertial characteristics compared to maritime and ground domains, combining relatively small steady accelerations with the widest vibration bandwidth. As summarized in Table~\ref{tab:domain_motion_characteristics}, commercial fixed-wing aircraft under normal climb and descent experience typical linear accelerations of approximately $0.01$--$0.03$\,m/s$^2$ in the vertical axis and $0.1$--$0.3$\,m/s$^2$ in the longitudinal axis, with dominant low-frequency components during ascent and descent~\cite{elmaghraby2020normal}. 
During takeoff, pitch angles at main-wheel lift-off for an Airbus A319 range from approximately $4.48^\circ$ to $12.4^\circ$, reflecting transient but significant attitude changes~\cite{zhang2025research}. 
In rotorcraft, hover under strong turbulent conditions can induce pitch variations from $0^\circ$ to $7^\circ$ and roll from $-15^\circ$ to $4^\circ$ over a $30$\,s interval~\cite{watson2025potential}. 
During cruise, aircraft vibration includes low-frequency components associated with turbulence (below 10\,Hz) and high-frequency components associated with engines (up to 1000\,Hz)~\cite{vibration5010007}.
Compared to maritime platforms (typically below $\sim$$1$\,Hz) and ground vehicles (dominant below $\sim$$25$\,Hz), aerospace environments therefore exhibit the widest frequency spectrum and the highest-frequency excitations.

For legged robots operating on aerial platforms, these characteristics introduce rapidly varying inertial conditions and high-frequency vibration disturbances. While steady accelerations may be moderate, transient pitch and roll excursions modify the effective gravity direction and support geometry, particularly during takeoff or rotorcraft hover. More critically, broadband vibrations that extend to hundreds of hertz can degrade state estimation, foot contact sensing, and control bandwidth. 
Reliable aerial deployment therefore demands disturbance-aware control, high-bandwidth state estimation, and robust perception under rapidly varying platform motions.

In summary, the main distinction across these application domains lies in the amplitude, bandwidth, and coupling structure of platform motion. These characteristics directly determine the disturbances experienced by the robot and the resulting difficulty of locomotion.

\begin{figure*}
    \centering
    \includegraphics[width=1\linewidth]{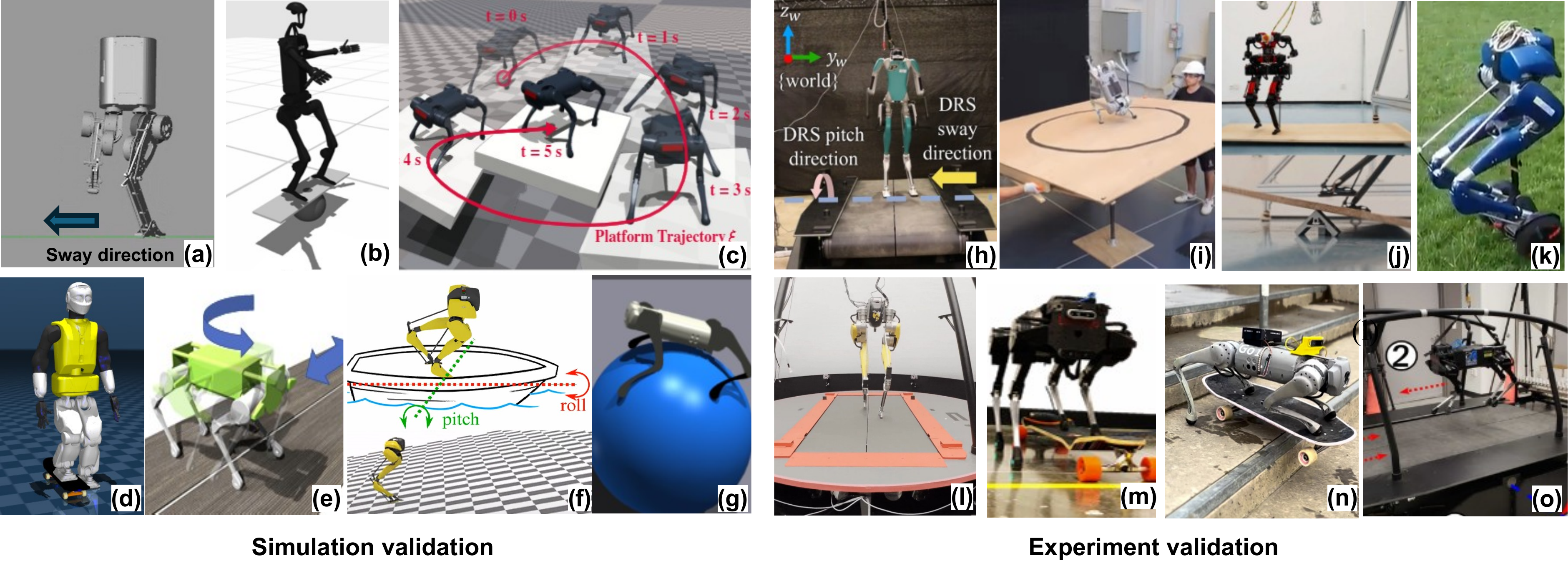}
    \vspace{-0.2 in}
    \caption{Screenshots of different existing studies related to non-inertial environments. Left: Studies with simulation validation, with images in (a)--(g) extracted from \cite{chen_contingency_2024},
    % (b)
    \cite{sferrazza_humanoidbench_2024},
    % (c)
    \cite{yoon_learning-based_2024},
    % (d)
    \cite{thibault2024learning},
    % (e)
    \cite{da2021learning},
    % (f)
    \cite{gu_robust-locomotion-by-logic_2025},
    % (g)
    \cite{ma_dreureka_2024}, respectively. Right: Studies with experimental validation, with images in (h)--(o) extracted from 
    \cite{gao_time-varying_2023},
    % (i)
    \cite{zinage2026contractionppo},
    % (j)
    \cite{luo_dcm-based_2025},
    % (k)
    \cite{gong_zero_2022},
    % (l)
    \cite{gu_robust-locomotion-by-logic_2025},
    % (m)
    \cite{da2021learning}, 
    % (n)
    \cite{liu_discrete-time_2025}, 
    % (o)
    and \cite{iqbal_provably_2020}, respectively.}
    \label{fig: existing study}
\end{figure*}

\section{Existing Modeling, State Estimation, and Control of Locomotion in Non-inertial Environments}

\label{sec:technical foundations}

Locomotion in non-inertial environments fundamentally differs from locomotion on stationary ground because the robot no longer moves on a fixed support, but within a contact-coupled robot-environment system. 
This shift in dynamics causes the degraded tracking performance and reduced stability margins of Boston Dynamics’ Spot, Ghost Robotics’ Vision 60, and Agility Robotics’ Digit humanoid robot on a dynamically moving platform~\cite{misenti_experimental_2025, gao_time-varying_2024}.

Non-inertial platform motion introduces persistent time-varying disturbances, altered contact conditions, and, for lightweight platforms, strong robot-environment coupling, thereby breaking the stationary-ground assumptions underlying many existing models \cite{kajita_3d_2001, xiong_3-d_2022, mesesan2023unified}, state estimators \cite{bloesch2013state, hartley2020contact, wisth2022vilens}, and controllers \cite{caron2019capturability, xiong_3-d_2022, mesesan2023unified, li_reinforcement_2025}. 
Moreover, platform motion is often only partially observable and difficult to predict, which can limit the robot’s ability to estimate or anticipate changes in support conditions.
This section reviews the existing modeling, estimation, and control approaches developed to address these non-inertial challenges, highlighting their key ideas, capabilities, and open problems.

\subsection{Modeling}

\subsubsection{Overview}

\begin{figure}
    \centering
    \includegraphics[width=0.95\linewidth]{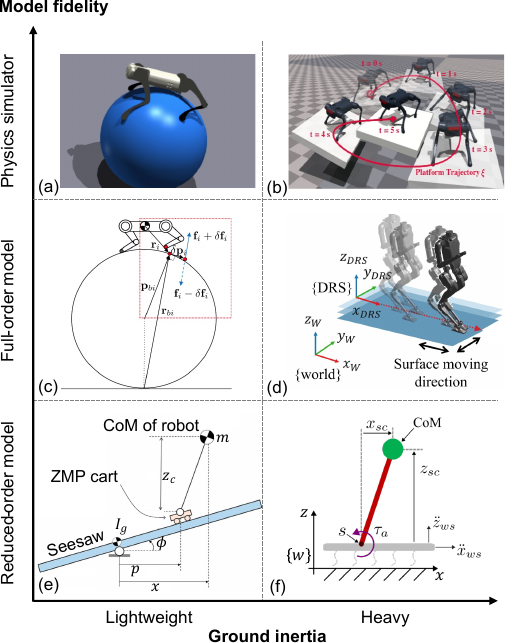}
    \vspace{-0.1 in}
    \caption{Representative existing legged robot models. 
    The images in (a)--(f) are extracted from \cite{ma_dreureka_2024}, \cite{yoon_learning-based_2024}, \cite{yang_dynamic_2020}, \cite{gao_time-varying_2024}, \cite{kimura_extended_2021}, and \cite{stewart_adaptive_nodate}, respectively.}
    \label{fig:4 modeling}
\end{figure}

Robot modeling seeks to represent system dynamics with sufficient fidelity and computational efficiency for real-time state estimation and control and offline training. In non-inertial environments, environment motion must be explicitly incorporated into the dynamics model (see Fig.~\ref{fig:4 modeling}) because it introduces inertial effects and time-varying disturbances that are absent in stationary-ground formulations.

A key distinction of non-inertial environments is whether environment motion is affected by the robot. For relatively lightweight platforms, such as small boats, robot and platform dynamics must be modeled jointly to capture bidirectional interaction. Although this preserves physical fidelity, it also increases the model dimensionality, underactuation, and partial observability. For large-inertia or rigidly actuated platforms, the environment can often be treated as an exogenous input. This avoids explicit bidirectional coupling, but turns otherwise time-invariant locomotion dynamics into time-varying systems with nonstationary constraints and disturbances.

\subsubsection{Full-order robot models}

Full-order models explicitly describe the dynamics for all degrees of freedom of the robot.
These models improve fidelity, but at substantially higher computational cost. 
For lightweight platforms, accurate prediction requires modeling both robot and platform dynamics together \cite{joshi_walking_2018}. For example, \cite{yang_dynamic_2020} modeled a quadruped walking on a yoga ball by coupling robot and ball dynamics through non-sliding contact constraints, while \cite{bouyarmane_quadratic_2019} augmented the system to include movable objects coupled to the robot through contact.

For heavy or rigidly actuated platforms, full-order models are often written as time-varying systems induced by time-varying holonomic constraints. On stationary rigid ground, contact constraints are generally trivial and time-invariant \cite{westervelt_feedback_2018}. In contrast, on moving platforms whose motion is assumed independent of the robot, non-inertial effects enter the model through time-varying contact constraints, where the contact surface motion is described as a time-varying trajectory \cite{iqbal_provably_2020}. This yields time-varying robot dynamics, and the main control challenge arises when the platform motion is unknown, partially known, or inaccurately estimated.

\subsubsection{Reduced-order robot models}

Reduced-order models retain the essential dynamics needed for planning and control at lower computational cost. On stationary ground, the linear inverted pendulum (LIP) is a standard abstraction of walking dynamics \cite{kajita_3d_2001}. In non-inertial environments, these models must also capture platform motion when robot-platform interaction is non-negligible~\cite{iqbal2021modeling}. For example, \cite{zheng_ball_2011} modeled a biped walking on a lightweight ball using an LIP on a massless wheel, yielding a linear time-invariant system that combines robot and platform states. Similar reduced-order models have been developed for balancing on a Bongo board \cite{nagarajan_balancing_2014}, Segway \cite{kimura_extended_2021, rajendran_humanoids_2024}, moving cart \cite{konishi_zmp_2023}, skateboard \cite{liu_discrete-time_2025}, and snakeboard \cite{anglingdarma_motion_2021}. A moving platform may also be represented as a vertically moving spring-mass-damper system \cite{asano_modeling_2021}.

For heavy or rigidly actuated platforms, reduced-order models remain low-dimensional but become time-varying because of environment motion. In LIP-based formulations, different directions of surface motion affect the robot dynamics differently \cite{stewart_adaptive_nodate}: vertical motion induces time-varying system parameters \cite{iqbal_analytical_2023, iqbal_asymptotic_2023, chen_contingency_2024, chen2024Locomotion}, whereas horizontal motion appears as a non-homogeneous disturbance \cite{gao_time-varying_2023}. Beyond the standard LIP, the angular-momentum LIP has also been extended to account for horizontal platform motion \cite{gao_time-varying_2024}. Although reduced-order models are typically attractive for real-time motion generation and control thanks to their computational efficiency, the time-varying effects induced by non-inertial motion remain challenging to control.

\subsubsection{Physics-based simulators}

Physics-based simulators are essential for developing and validating locomotion methods in non-inertial environments before hardware deployment. They also provide scalable training environments for learning-based controllers.
Compared with reduced- and full-order analytical models, they capture effects such as multi-contact, self-collision, and actuator dynamics that are often simplified or neglected.
Widely used simulators, including MuJoCo \cite{todorov_mujoco_2012}, Isaac Gym \cite{makoviychuk2isaac}, Isaac Lab~\cite{IsaacLab2024}, PyBullet \cite{e_coumans_and_y_bai_pybullet_2016}, and RaiSim \cite{hwangbo_per-contact_2018}, fairly accurately describe multi-rigid-body dynamics and robot-environment contact, and can be extended with moving platforms to study locomotion in non-inertial settings.

Yet, simulating non-inertial locomotion often requires augmenting standard physics engines with environment-specific motion models. For example, maritime locomotion may be studied using simulators such as VRX \cite{bingham_toward_2019}, which provided wave-driven vessel motion relevant to shipboard operation. Such tools extend general-purpose simulators by incorporating the characteristic dynamics of the surrounding environment.

\input{table/model_table}

\subsubsection{Open problems}

Despite recent progress, analytical (reduced- or full-order) models for non-inertial locomotion remain limited by strong assumptions. For lightweight platforms, most studies consider balancing or walking on platforms with limited DoF, leaving bipedal locomotion on six-DoF platforms largely unexplored. For heavy platforms, most reduced-order models rely on LIP-like, point-foot formulations and often neglect ankle torque, platform pitching and rolling, and foot slippage, even though these factors can strongly affect contact timing, body orientation, and stability. Relaxing such assumptions improves generality but typically increases model complexity and challenges the design of real-time estimators and controllers based on these models.

One direction for overcoming these limitations is to augment or replace simplified analytical models with data-driven models. Hybrid approaches that combine analytical structure with learned residual corrections have been effective for legged systems with soft legs \cite{yang_legged_2021} and complex actuators \cite{hwangbo2019learning}, and may be extended to non-inertial platform dynamics driven by disturbances such as waves and wind. More expressive alternatives, such as learned world models conditioned on state and action \cite{wu2023daydreamer}, may further improve prediction in highly uncertain environments.

Another open question is the choice of reference frame. Most existing approaches inherit inertial-frame formulations from stationary-ground locomotion. However, in non-inertial environments, robot motion relative to the moving platform may be more relevant for control, while inertial-frame pose estimation may be difficult in practice. Conversely, platform-attached formulations introduce additional complexity, including time-varying effective gravity and fictitious forces. Understanding this trade-off is important for developing models that remain both practical for estimation and robust in control.

\subsection{State Estimation}

\subsubsection{Overview}
State estimation aims to recover quantities that characterize a robot's motion over time, including position, velocity, orientation, and sometimes contact states or joint angles~\cite{barfoot2024state}. In inertial environments, this is typically done with respect to a fixed inertial reference frame under the assumption of negligible environment acceleration. Under these conditions, classical frameworks such as visual-inertial odometry~\cite{geneva2020openvins, qin2018vins} and simultaneous localization and mapping (SLAM)~\cite{rosen2021advances, slam-handbook} have achieved strong performance in wheeled and aerial robots.

For legged robots, state estimation is highly challenging because of intermittent contact and changing terrain. Leg inertial odometry addresses this by fusing inertial, kinematic, and contact information~\cite{bloesch2013state, hartley2020contact}, sometimes together with cameras or LiDAR to reduce drift in visually structured environments~\cite{wisth2022vilens}. These methods, however, generally assume rigid contact with a stationary support surface.

In non-inertial environments, as the support surface itself moves, classical contact assumptions no longer apply. This introduces coupled robot-platform motion and new observability challenges. Existing work in this setting can be viewed through three related problems: relative state estimation, which estimates robot motion with respect to the moving platform; absolute state estimation, which estimates robot motion with respect to a fixed inertial frame; and platform state estimation, which estimates the motion of the supporting surface itself.

\subsubsection{Relative state estimation}

For relative state estimation, \cite{he_legged_2024} extended the leg-inertial odometry to dynamic platforms by introducing an auxiliary inertial measurement unit (IMU) on the surface and using an Invariant Extended Kalman Filter (InEKF)~\cite{barrau2016invariant, barrau2018invariant} to estimate robot pose relative to the moving platform. A subsequent study~\cite{he_invariant_nodate} extended this idea to wheeled robots and analyzed observability, showing that the full relative state becomes observable when both robot and platform carry IMUs and the platform undergoes general dynamic motion.
Later on, this approach was further extended to be fully proprioceptive without relying on external ground motion sensing~\cite{mandali2026}.

Despite this progress, reliable relative estimation without direct inertial sensing of the platform motion remains open. One promising direction is to use robot-centric vision to infer relative pose between the robot and its support surface. Related formulations have been studied for drone landing and docking~\cite{semerikov2025vision, xin2022vision}, but extending them to legged robots is nontrivial because
the measurements from the robot's onboard IMUs
reflect coupled robot-platform motion rather than robot motion alone.

\subsubsection{Absolute state estimation}
For absolute state estimation, the goal is to recover the robot pose with respect to a fixed Earth frame. When environmental motion is weak or short-lived, it may be treated as a disturbance within the estimator~\cite{yu2023fully, bloesch2013state}. However, this becomes inadequate when the support surface undergoes sustained or large-amplitude motion. To address this, \cite{zhang_pose_2016, zhang2018absolute} proposed a dual-gyroscope method for absolute attitude estimation using one sensor on the robot and one on the platform, fused in an Extended Kalman Filter (EKF) through coupled kinematic constraints. Related ideas were explored in human-bicycle systems using additional force sensing and multiple-EKF formulations \cite{zhang2013rider, zhang_whole-body_2016}. For legged robots, \cite{gao2021invariant, gao_invariant_2022} developed a right-InEKF using a body-mounted IMU, leg encoders, and on-platform fiducial markers. The method accounts for nonstationary contact points and hybrid locomotion, and was validated on a pitching treadmill. However, it assumes full knowledge of the platform's six-dimensional (6-D) pose in the Earth frame.

\subsubsection{Platform state estimation}

Platform state estimation is also important because support-surface motion directly affects robot dynamics and whole-body control. For locomotion in non-inertial environments, not all platform states are equally relevant: trunk roll and pitch, angular velocity, and linear acceleration often matter more than global position or yaw, since they directly affect effective gravity and inertial loading felt by the robot. A systematic study of which platform states are most relevant for locomotion, however, remains lacking.

The maturity of platform-state estimation also differs by domain. Aircraft and ground vehicles benefit from decades of work on filtering-based inertial navigation systems~\cite{lefferts1982kalman, madyastha2011extended},
integration with the global navigation satellite system~\cite{sukkarieh2002high}, and SLAM-style perception~\cite{zou2021comparative, bresson2017simultaneous}. In contrast, surface vessels~\cite{qiao2023survey, hitz2015relaxing} and trains remain less studied, particularly for short-horizon estimators tailored to onboard locomotion.

\begin{table*}[t]
\centering
\caption{Representative state estimation methods for robots operating in non-inertial environments.}
\label{tab:survey_state_estimation}
\footnotesize
\begin{tabular}{p{3.0cm} p{3.2cm} p{1.4cm} p{5.0cm} p{3.3cm}}
\toprule
\textbf{Estimated variables} & \textbf{Sensors} & \textbf{Method} & \textbf{Key assumption} & \textbf{Performance focus} \\
\midrule

Relative robot pose~\cite{he_legged_2024}
& Robot IMU, leg kinematics/contact, platform IMU 
& InEKF 
& Requires an auxiliary IMU mounted on the platform 
& Observability, convergence rate, accuracy \\

Relative robot pose~\cite{he_invariant_nodate} 
& Robot IMU, encoders, platform IMU 
& InEKF 
& Full relative state becomes observable under general dual-IMU dynamic motion 
& Observability, convergence rate, accuracy \\

Absolute human attitude~\cite{zhang_pose_2016,zhang2018absolute} 
& Body gyroscope, platform gyroscope 
& EKF 
& Relies on a dual-gyroscope setup 
& Observability, accuracy \\

Coupled human-platform pose~\cite{zhang2013rider,zhang_whole-body_2016} 
& IMU/gyroscope + force sensing 
& EKF/multi-EKF 
& Developed for rider-bicycle systems rather than legged robots 
& Accuracy \\

Absolute robot pose~\cite{gao_invariant_2022} 
& Body IMU, leg encoders, platform fiducial markers 
& InEKF 
& Assumes the platform motion is accurately known or estimated 
& Observability, convergence rate, accuracy, robustness \\

\bottomrule
\end{tabular}
\end{table*}

 \subsubsection{Open problems}
Despite recent progress, state estimation in non-inertial environments remains largely underexplored. Several open problems continue to limit both theoretical understanding and practical deployment.

First, a central question is which estimated quantities are actually needed for locomotion control and planning. On stationary terrain, world models, such as scene graphs~\cite{rosinol20203d, rosinol2021kimera} or semantic maps~\cite{gan2020bayesian, kostavelis2015semantic}, are often expressed in a unified gravity-aligned inertial frame, which simplifies control and planning. On dynamic platforms, however, the relationship between estimation accuracy and locomotion performance becomes less clear. Determining which absolute and relative states, such as platform-relative pose, velocity, or contact states, are sufficient for stability and task execution remains largely uncharacterized.

Second, a general understanding of when absolute and relative states are observable in non-inertial settings is still lacking. Existing analyses are often limited to specific sensing configurations~\cite{zhang2018absolute} or filtering formulations~\cite{he_legged_2024, he_invariant_nodate}. Broader observability studies across sensing modalities and operational conditions could guide both sensor-suite design and estimator architecture. For example, it remains unclear under what visual and inertial conditions the relevant states become observable for a legged robot on a moving platform.

Third, model-based recursive filters such as EKF and InEKF remain the dominant paradigm, but they have limited ability to represent long-horizon correlations and complex hybrid dynamics. Factor graphs~\cite{dellaert2012factor, slam-handbook, wisth2019robust} offer a natural extension for enforcing long-horizon consistency through loop closures and multi-sensor fusion. In parallel, learning-based methods may help capture nonlinear robot-platform interactions, building on recent progress in learned state estimation~\cite{buchanan2022learning, youm2025legged} and robot dynamic modeling~\cite{skulstad2020hybrid, li2025robotic}.

Fourth, progress is also limited by the lack of public datasets and standardized benchmarks for non-inertial locomotion. Most existing datasets~\cite{patel2025tartanground, Tuna-Frey-Fu-RSS-25, 10631676} assume stationary terrain and do not capture coupled robot-platform dynamics in settings such as ships, aircraft, or trains. Moreover, current evaluations are often restricted to controlled treadmill experiments. Open datasets, high-fidelity simulation environments, and standardized benchmarks would enable more reproducible comparison and accelerate the development of robust and generalizable estimators.

\subsection{Control and Planning}
  
\subsubsection{Overview}

Control and planning for legged robots can be viewed as a three-level hierarchy. High-level planning generates task-space objectives such as center-of-mass, swing-foot, or end-effector trajectories. Middle-level motion generation converts these objectives into contact-consistent and dynamically feasible joint-space references, accounting for gait phases, contact states, and timing constraints. Low-level control then computes motor commands, such as joint torques, to track these references while satisfying full-body dynamics and physical constraints.

In non-inertial environments, all three levels must account for platform motion. Time-varying accelerations and the associated inertial effects alter viable footsteps, support conditions, and locomotion stability. Thus, planners should predict locomotion under moving support surfaces, while controllers must track desired motion and treat additional inertial loading.

From a control perspective, one important distinction is the assumed knowledge of platform motion. Existing methods can be broadly grouped into three regimes: controllers that exploit known or measured platform motion, controllers that assume only bounded disturbances and seek robust feasibility, and controllers that must handle genuinely unknown, multidirectional, and possibly non-periodic motion. This distinction determines whether the controller can plan predictively or must instead react conservatively to uncertainty.
Within this broader perspective, existing approaches can still be categorized by controller architecture~\cite{gu2025evolution}, including classical feedback control, optimization-based control, and learning-based control (see Table \ref{tab:control}).

\subsubsection{Classical feedback control}
Classical feedback control methods aim to stabilize locomotion, reject disturbances, and attenuate modeling errors through reactive feedback.
Representative classical approaches use reduced-order models such as angular-momentum LIP (ALIP) \cite{gong_zero_2022} to generate foot placement for stable walking. These low-dimensional models enable efficient planning, while the resulting references are executed by full-order whole-body controllers, such as quadratic-program-based controllers \cite{bouyarmane_quadratic_2019} that enforce kinematic consistency and contact constraints. Related approaches based on divergent component of motion (DCM) \cite{khadiv2020walking} and capture points \cite{caron2019capturability} followed the same principle. However, most were developed for inertial settings and do not explicitly account for moving support surfaces.

A foundational step toward feedback control in non-inertial environments was made by \cite{iqbal_provably_2020}, which introduced one of the earliest provably stabilizing control frameworks for legged locomotion under platform motion. By modeling robot-ground interaction as a hybrid time-varying system and drawing upon Lyapunov stability theory, the method explicitly incorporates platform motion into the dynamics model and compensates for it through feedback control. This work also demonstrated one of the first experimental validations of legged locomotion on a dynamically pitching platform.

Building on this idea, \cite{gao_time-varying_2024} extended ALIP to include swaying surface velocity, enabling disturbance-aware motion generation of footsteps and momentum targets. Still, the method assumes periodic, known, single-direction sway. To relax such assumptions, \cite{stewart_adaptive_nodate} proposed an adaptive controller with ankle-torque adaptation to handle unknown, multidirectional, and potentially aperiodic platform motion.

Despite these advances, classical feedback methods do not explicitly optimize performance objectives or systematically enforce state and input constraints. This motivates optimization-based control.

\subsubsection{Optimization-based control}

Optimization-based control formulates locomotion as a finite-horizon optimal control problem, computing control actions that are stabilizing, performance-aware, and constraint-consistent.
Optimization-based methods, including convex model predictive control (MPC), nonlinear MPC, and trajectory optimization, have been highly successful in inertial environments. Convex MPC on the MIT Cheetah 3 quadruped \cite{di_carlo_dynamic_2018} optimized ground reaction forces in under 1~ms and enabled dynamic gaits such as trotting, galloping, and pronking. Nonlinear MPC can handle more complex dynamics and contacts. For example, \cite{bellicoso_dynamic_2018} used it on the ANYmal quadruped to optimize centroidal dynamics with explicit contact constraints for agile locomotion. Longer-horizon trajectory optimization has also enabled highly dynamic motions such as jumping and obstacle clearance \cite{winkler2018gait}. Most of these methods, however, assume inertial reference frames and stationary ground.

Recent work has begun to extend optimization-based control to non-inertial environments. Existing approaches generally fall into three categories: (i) robust or contingency MPC that models platform motion as a bounded disturbance and guarantees feasibility under worst-case acceleration or jerk bounds~\cite{chen_contingency_2024,chen2024Locomotion}; (ii) embedding simplified platform dynamics (e.g., vertical oscillations) into reduced-order models such as hybrid time-varying LIP (HT-LIP), enabling footstep optimization via quadratic programming under measured or estimated vertical motion~\cite{iqbal_ht-lip_nodate}; and (iii) treating ground motion as an external perturbation and synthesizing recovery trajectories subject to safety constraints~\cite{gu_robust-locomotion-by-logic_2025}.

These methods mark important progress, but most still assume partially known motion, simplified oscillation patterns, or disturbance bounds that do not exploit ground motion structure. They also rarely account for characteristic properties of real transportation environments, such as low-frequency maritime oscillations, moderate-frequency vehicle vibrations with transient accelerations, or broadband aerospace disturbances. In addition, many optimization-based methods rely on nominal gait schedules. Under platform motion, unmodeled changes in support conditions can shift touchdown timing, degrade tracking, and introduce destabilizing impacts. Estimation errors in platform motion or non-inertial forces can further degrade prediction accuracy. These issues motivate controllers that can adapt online to unknown, non-periodic, and multidirectional disturbances.

\input{table/control_table}

\subsubsection{Reinforcement learning}

Reinforcement learning (RL) provides a data-driven alternative by formulating locomotion control as a sequential decision-making problem, where a policy is trained to map observations to control actions to maximize a long-term reward. By learning directly from interaction (often through physics-based simulations), RL can handle complex dynamics and environments where analytical models are insufficient. RL has progressed from simulation-only gait learning to sim-to-real transfer via domain randomization, and further to hierarchical control and perception-integrated multi-skill locomotion.

Recent RL-based locomotion systems have shown strong robustness in inertial environments~\cite{tay2026hybridmimic}. For example, \cite{li_reinforcement_2025} trained bipedal policies for walking, running, and jumping across diverse terrains, while \cite{hoeller_anymal_2024} combined perception-driven scene reconstruction with hierarchical RL to enable parkour-like navigation in ANYmal. However, these methods generally assume stationary ground without considering persistent platform-induced accelerations.

More recent work begins to incorporate non-inertial effects into learning. Compared with robust MPC, RL can internalize temporal disturbance patterns and adapt through experience. LAS-MP \cite{yoon_learning-based_2024} jointly trained a balancing policy and state estimators through curriculum learning over 6-DoF platform motion. DHAL \cite{liu_discrete-time_2025} integrated hybrid mode identification into RL to improve robustness across contact transitions and enabled behaviors such as quadruped skateboarding. DrEureka \cite{ma_dreureka_2024} improved sim-to-real transfer by automating reward shaping and domain randomization for balancing on dynamically unstable objects.

\subsubsection{Open problems}
Despite progress in classical feedback, optimization-based, and RL-based control, several fundamental challenges remain.

First, the unknown disturbances induced by ground motion remain challenging as the disturbances may change rapidly in magnitude.
While robust and tube-based MPC can enhance robustness by assuming constant or bounded disturbances, they face a sharp feasibility-robustness trade-off.
Larger disturbance sets increase robustness but are also too conservative, which narrows the feasible region and may render the optimization problem infeasible.
Promising directions include adaptive disturbance-set tuning based on estimator confidence, probabilistic MPC \cite{6761117} that enforces constraints probabilistically rather than worst-case, and hybrid predictors that combine physics-based models and learned residuals to shrink uncertainty online.

Second, rigid contact assumptions break down on moving platforms.
Analytical-model-based controllers
often enforce sticking contact by imposing zero relative motion at the foot-ground contact interface.
However, foot slip and rolling violate this assumption and can cause infeasible force solutions or degraded tracking. 
To address these complex contact conditions, future model-based controllers can benefit from integrating real-time contact-state estimation, predictive slip detection, and hybrid or complementarity-based optimization to handle dynamic contact transitions consistently.

Third, RL handles unmodeled dynamics but lacks structure and guarantees.
RL has shown strong empirical adaptation to diverse complex terrains, but most policies are still trained under inertial assumptions and treat platform motion as randomized disturbance. This limits generalization to structured, significant non-inertial motion. Hierarchical approaches such as \cite{zhang_motion_2025}, which combine motion priors with residual adaptation, suggest a promising direction: inertial locomotion policies could provide structured priors, while residual learning captures platform-specific effects. Still, RL lacks formal guarantees and explicit treatment of robot-platform coupling.

\section{Research Challenges and Future Directions}
\label{sec:research challenges and future}

Beyond modeling, state estimation, and control, several broader challenges still limit useful robot operation in non-inertial environments, such as autonomy, system-level design, bio-inspiration, versatile functionality, safety, and testing. This section highlights these challenges and outlines promising research directions.

\subsection{Autonomy}

Full autonomy in non-inertial environments remains largely unexplored. Achieving autonomy requires tight integration of perception, high-level planning, and low-level control, since persistent platform motion can propagate failures across these layers. Recent advances in computer vision (CV) and natural language processing (NLP) may further support higher-level reasoning and adaptation in such settings. The following subsections discuss perception, planning, and related emerging tools for enabling autonomy in non-inertial environments.

\subsubsection{Perception}

Perception in autonomous systems extends beyond state estimation by providing geometric and semantic understanding of the environment. In stationary settings, this often means mapping terrain and identifying stable footholds \cite{ bertrand2020detecting}. In non-inertial environments, perception should additionally separate robot ego-motion and obstacle motion from exogenous platform motion \cite{wang2024survey}. Beyond estimating the current state, perception should also help anticipate future disturbances from environmental cues. For example, a robot on a tilted vessel may need to infer impending motion from visual or auditory observations in order to plan safely.

\subsubsection{High-level task and motion planning}

High-level planning translates estimated states and environmental understanding into structured high-level actions. It can be divided into task planning and motion planning \cite{guo2023recent}. Task planning determines the sequence of operations needed to achieve a goal. In non-inertial environments, this sequence may need to adapt to platform motion; for example, a robot may postpone a high-risk maneuver until the disturbance becomes smaller. Motion planning then generates feasible task-space or joint-space trajectories. Under non-inertial conditions, these plans must remain feasible in the presence of external disturbances while respecting physical constraints such as friction limits and actuation bounds.

\subsubsection{Emerging tools from CV and NLP}
World models and Vision-Language-Action (VLA) models may enhance autonomy in non-inertial environments through higher-level reasoning and adaptation. World models learn predictive representations of environment dynamics \cite{wu2023daydreamer, hafner2023mastering} and could help capture platform motions that are difficult to model analytically, such as ship motion under varying sea states. VLA models map multimodal inputs (e.g., vision and language) to robot actions \cite{black2410pi0, zitkovich2023rt}, and may help robots adapt behavior using cues about motion patterns and task context.

\subsection{System-level Design}

Robot performance in non-inertial settings can be substantially improved through system-level design that integrates hardware and algorithms. Advances in sensors, actuators, and mechanisms can collectively improve robustness to persistent disturbances and rapidly changing dynamics.

\subsubsection{Sensors}

Reliable sensing is essential for perception and state estimation in non-inertial environments, where persistent ground motion can degrade conventional estimators. Vision can help correct long-term IMU drift \cite{wisth2022vilens}, but rapid motion may introduce blur and reduce performance. Event cameras provide a promising alternative because they offer high-speed, low-latency, and high-dynamic-range sensing without motion blur \cite{gallego_event-based_2022}. Their recent use in tracking highly dynamic objects \cite{da_costa_new_2025} and event-based visual odometry \cite{chen_esvio_2023} suggests strong potential for legged robot operation on moving platforms.

Additionally, time-varying platform motion can make contact timing and location uncertain. Robust tactile and force sensors can provide whole-body contact information \cite{luo_tactile_2025, murooka_whole-body_2024, zhou2026tactile}, which may improve state estimation and control under unpredictable support motion.

\subsubsection{Actuators}

Actuator design is critical for locomotion in non-inertial environments, where robots often need to respond to persistent disturbances and unexpected impacts. These settings favor actuators that combine high power density, fast torque response, and mechanical compliance. Quasi-direct-drive actuators provide high torque bandwidth and backdrivability for rapid disturbance rejection \cite{wensing_proprioceptive_2017}, while series elastic actuators \cite{pratt_series_2004} and variable stiffness actuators \cite{liu_impedance-controlled_2020} help absorb shocks and tolerate uncertain contact. Another promising direction is to incorporate energy storage and recovery into actuator design, which may increase efficiency in oscillatory environments such as wave-driven vessels.

\subsubsection{Mechanism design}

% \cite{wang_mechanical_2023}

Mechanism design helps mitigate the perception and control challenges of locomotion in non-inertial environments. Legged locomotion typically relies on non-prehensile contact with the ground and may fail when ground motion narrows feasible contact conditions, for example through limited friction margins. Alternative foot designs can expand the feasible contact force space and reduce the burden on control. Examples include electromagnetic feet for adhesion on vertical or inverted surfaces \cite{noauthor_agile_nodate}, clawed mechanisms that secure contact during dynamic transitions \cite{yim2025monopedal}, and jammed granular feet that adapt contact to the surface \cite{narayanan2025adaptive}.
Additional actuation can further improve robustness to non-inertial motion. For example, propellers or flapping wings have been used to enhance stability on challenging terrain \cite{kim2021bipedal} and to help hopping robots remain stable on moving platforms \cite{hsiao2025hybrid}. More broadly, such designs move legged locomotion beyond purely frictional, non-prehensile interaction toward prehensile contact and hybrid actuation, offering additional means to counteract platform-induced disturbances.

\subsection{Bio-inspiration}

Animals exhibit remarkable physical capabilities in complex environments, and their strategies have long inspired robot perception and control \cite{Ijspeert_Biorobotics_2014}. In legged locomotion on stationary ground, for example, humans rely on ankle, hip, and stepping strategies to reject disturbances \cite{winter_human_1995}, which in turn have inspired related stabilization mechanisms in legged robots \cite{jeong_robust_2019}. Yet, while recent robotics research has increasingly emphasized advanced computational methods for locomotion, human locomotion remains an underexplored source of insight for control in complex and non-inertial environments.

This gap is important because walking in non-inertial environments is challenging even for humans \cite{kao_motor_2022}, and human adaptation in such settings reveals principles that may be valuable for robotics. 
On moving platforms, human gait characteristics also depend on the disturbance properties: step timing varies with walking direction on a floating vessel \cite{haaland_human_2015}, and walking speed is more strongly affected by platform rotation amplitude than by motion period \cite{guo_experimental_2025}. During stance on harmonically moving platforms, humans further exhibit frequency-dependent balance adaptation: at low frequencies they rely primarily on ankle strategies and favor energy efficiency, whereas at higher frequencies they shift toward hip strategies and prioritize stability \cite{taleshi_humans_2025, lu_switched_2025}. These findings suggest that disturbance-dependent strategy switching, disturbance estimation and compensation, and multisensory feedback provide useful principles for robust robot locomotion in non-inertial environments, yet these mechanisms remain largely unexplored in current robot perception and control.

Besides control, perception and estimation of disturbance are also essential for human balancing in non-inertial environments.
To achieve postural control on moving platforms, humans utilize the integrated information from multiple sensing systems, including vision, tactile, and proprioception \cite{Horak_Macpherson_1996}, to estimate and compensate for the disturbance \cite{Lippi_Bioinspired_2021}.

\subsection{Versatile Functionality}

Beyond basic locomotion, useful deployment in non-inertial environments requires legged robots to perform tasks while maintaining stability under platform motion. The central challenge is therefore not locomotion alone, but diverse task execution under moving support involving manipulation, loco-manipulation, and human-assistive applications.

\subsubsection{Manipulation and loco-manipulation}

Manipulation on a moving support is challenging because platform motion affects end-effector behavior~\cite{donald2024adaptive}. Even simple positioning tasks can become difficult when oscillation or acceleration introduces fictitious forces that alter the effective dynamics of the robot and manipulated object. Early work~\cite{5152666} on manipulators mounted on ships and offshore structures indicated that explicitly incorporating platform motion into dynamics, planning, and control can help ensure reliable performance. More recent work \cite{9425004} used predictive end-effector control and base-relative trajectory tracking to compensate for platform motion. For legged robots, the challenge is broader: manipulation and locomotion must be coordinated simultaneously, since the robot must regulate both balance and task execution under the same time-varying disturbance. 

\subsubsection{Human-assistive and human-robot systems}

Human-assistive robots, such as exoskeletons and prostheses, face an additional requirement in non-inertial environments: they must support not only task execution but also human stability. On a moving bus, ship, or industrial platform, the device should help the user maintain footing while adapting to platform motion in real time. This calls for user-in-the-loop models that predict human balance responses under external disturbances, adaptive impedance control that accounts for user intent and fatigue, and multi-sensor fusion that combines body-worn sensing with platform-motion estimation. A long-term goal is personalized and learning-mediated assistance that co-adapts with the user across different moving environments.

\subsubsection{Generalization across non-inertial environments}

A broader open challenge is to generalize these capabilities across diverse non-inertial environments without task-specific reprogramming. Moving rigid supports, such as ships or trucks, and deformable or flowing substrates, such as inflatable boats or mud-like terrain, introduce different but overlapping challenges. A practical robot should not require a separate controller for each case. Promising directions include modular control architectures that separate predictable and unpredictable disturbance components, as well as meta-learning or adaptation frameworks that transfer stability and task skills across platforms with different dynamics.

\subsection{Safety}

Most existing safe control methods, including control Lyapunov functions \cite{anand_safe_2021}, control barrier functions \cite{agrawal_discrete_2017}, constrained optimization \cite{compton_dynamic_2025}, and probabilistic risk-aware strategies \cite{akella_risk-aware_2024}, are developed mainly for continuous nonlinear systems. Extending them to legged locomotion in non-inertial environments is difficult because the dynamics are hybrid, high-dimensional, and time-varying, and the relevant safety constraints may also change with platform motion \cite{brunke_safe_2022}.

Data-driven methods may help address this gap~\cite{kim2026modular}. For example, physics-informed neural networks can be used to learn safety or stability certificates for high-dimensional legged systems \cite{bansal2021deepreach}, and latent-space methods may offer a lower-dimensional representation for safety reasoning \cite{agrawal2025anysafe}. In non-inertial environments, however, the main challenge is generalization across diverse platform motions. Promising directions include training under broad motion randomization, identifying platform motion from limited onboard sensing through adaptation or distillation \cite{lee2020learning}, and incorporating inductive bias from platform dynamics, such as equivariant structures \cite{LieNeuron}.

\subsection{Robot Testing}
Systematic testing for legged locomotion in non-inertial environments remains underdeveloped, limiting fair comparison across methods and slowing real-world deployment. In inertial settings, more structured evaluation is beginning to emerge through repeatable physical tests \cite{weng_towards_2024}, standardized benchmarking procedures \cite{weng_comparability_2023}, and simulation-based platforms \cite{ zhang_generating_2023}. By contrast, evaluation in non-inertial environments is still largely based on author-defined and limited platform motions \cite{gao_time-varying_2024, gu_robust-locomotion-by-logic_2025, bermudez2025comparative}, with \cite{misenti_experimental_2025} standing out as a rare real-world vessel study.
Standardized testing protocols and evaluation metrics for non-inertial locomotion therefore remain immature. Even basic questions, such as how to define stability or which reference frame should be used to measure tracking error on an accelerating platform, are still unresolved. The field would benefit from systematic and reproducible testing frameworks, in both simulations and experiments, to enable fair comparison, guide controller development, and support deployment in real-world non-inertial settings.

\section{Conclusion}
\label{sec:conclusion}
This survey has reviewed the state of the art in modeling, state estimation, and control for legged locomotion in non-inertial environments. The existing literature shows that accelerating support surfaces introduce persistent time-varying disturbances, altered contact conditions, and coupled robot-environment dynamics that fundamentally challenge the stationary-ground assumptions underlying conventional legged locomotion methods. Although recent progress has demonstrated promising advances in modeling, state estimation, and control, important limitations remain in model generality, environment-state observability, control robustness, and experimental validation under realistic operating conditions. It is our hope that this survey will help clarify the key technical challenges, provide a useful reference for researchers entering this area, and stimulate further progress toward reliable legged robots capable of reliable and versatile operation in real-world non-inertial environments.

\bibliographystyle{Transactions-Bibliography/IEEEtran}
\bibliography{references}

\begin{IEEEbiography}[{\includegraphics[width=1in,height=1.15in,clip]{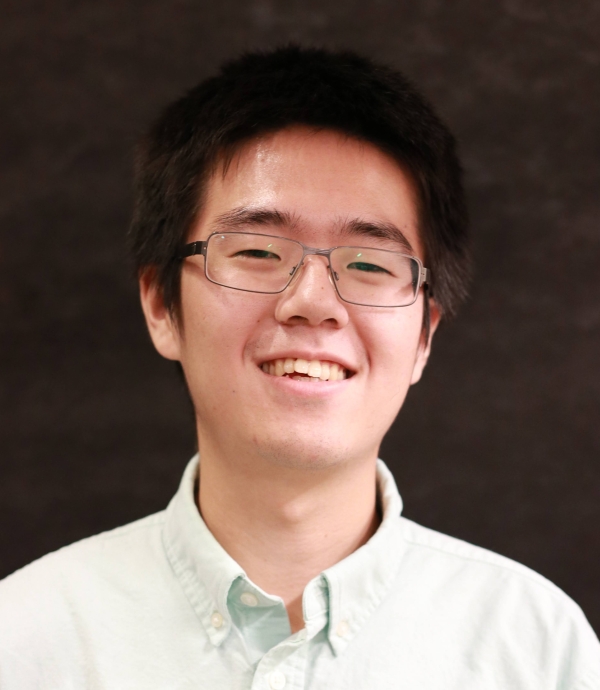}}]{I-Chia Chang} received the B.S. degree in mechanical engineering from National Taiwan University, Taipei, Taiwan, in 2022. He is currently working toward the Ph.D. degree in mechanical engineering with Purdue University, West Lafayette, IN, USA. He was a recipient of the Ross Fellowship from Purdue University. His research interests include motion planning and control for legged robots.
\end{IEEEbiography}

\vspace{-0.3 in}

\begin{IEEEbiography}[{\includegraphics[width=1in,height=1.15in,clip,keepaspectratio]{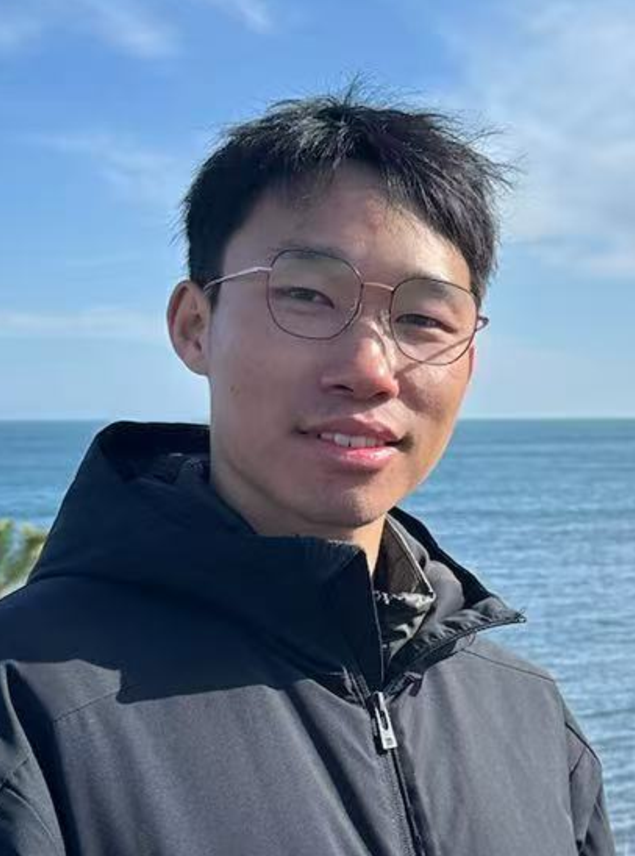}}]{Xinyan Huang} received the B.S. degree in mechanical engineering from Hefei University of Technology, Anhui, China, in 2020, and the M.Eng. degree in mechanical engineering from Zhejiang University, Hangzhou, China, in 2023. He is currently working towards the Ph.D. degree in mechanical and aerospace engineering with Rutgers University, Piscataway, NJ, USA.

His research interests include robotics, optimal bipedal motion planning, whole-body dynamics control, and mechatronics.
\end{IEEEbiography}

\vspace{-0.3 in}
\begin{IEEEbiography}
[{\includegraphics[width=1in,height=1.15in,clip]{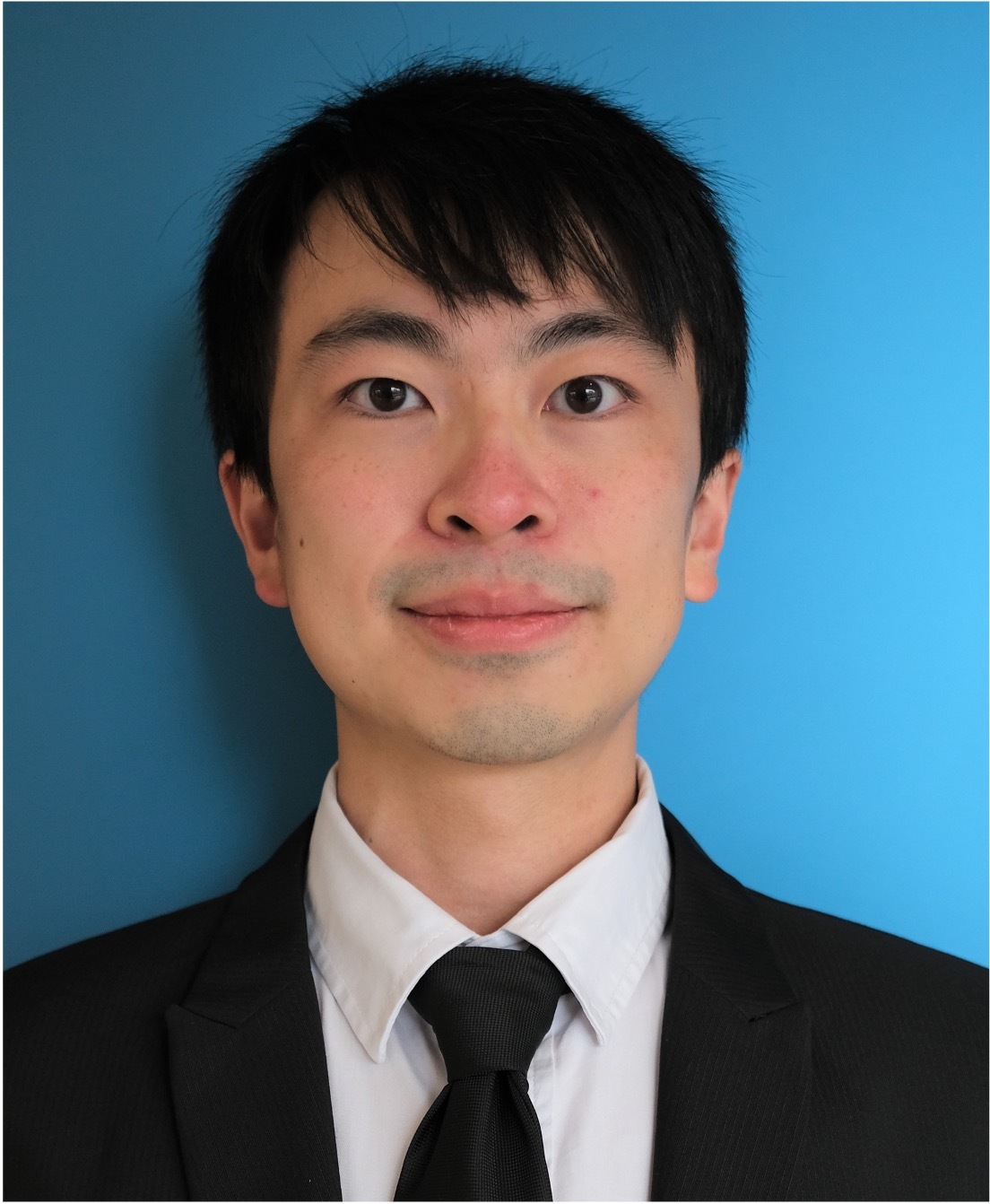}}] {Tzu-Yuan (Justin) Lin} received the B.S. degree in Mechanical Engineering from National Taiwan University in 2017, the M.S. degree in Robotics from the University of Michigan, Ann Arbor in 2020, and the Ph.D. degree in Robotics from the University of Michigan, Ann Arbor in 2024.
His research interests include robot perception, computer vision, and geometric deep learning.
\end{IEEEbiography}

\vspace{-0.3 in}
\begin{IEEEbiography}
[{\includegraphics[width=1in,height=1.15in,clip]{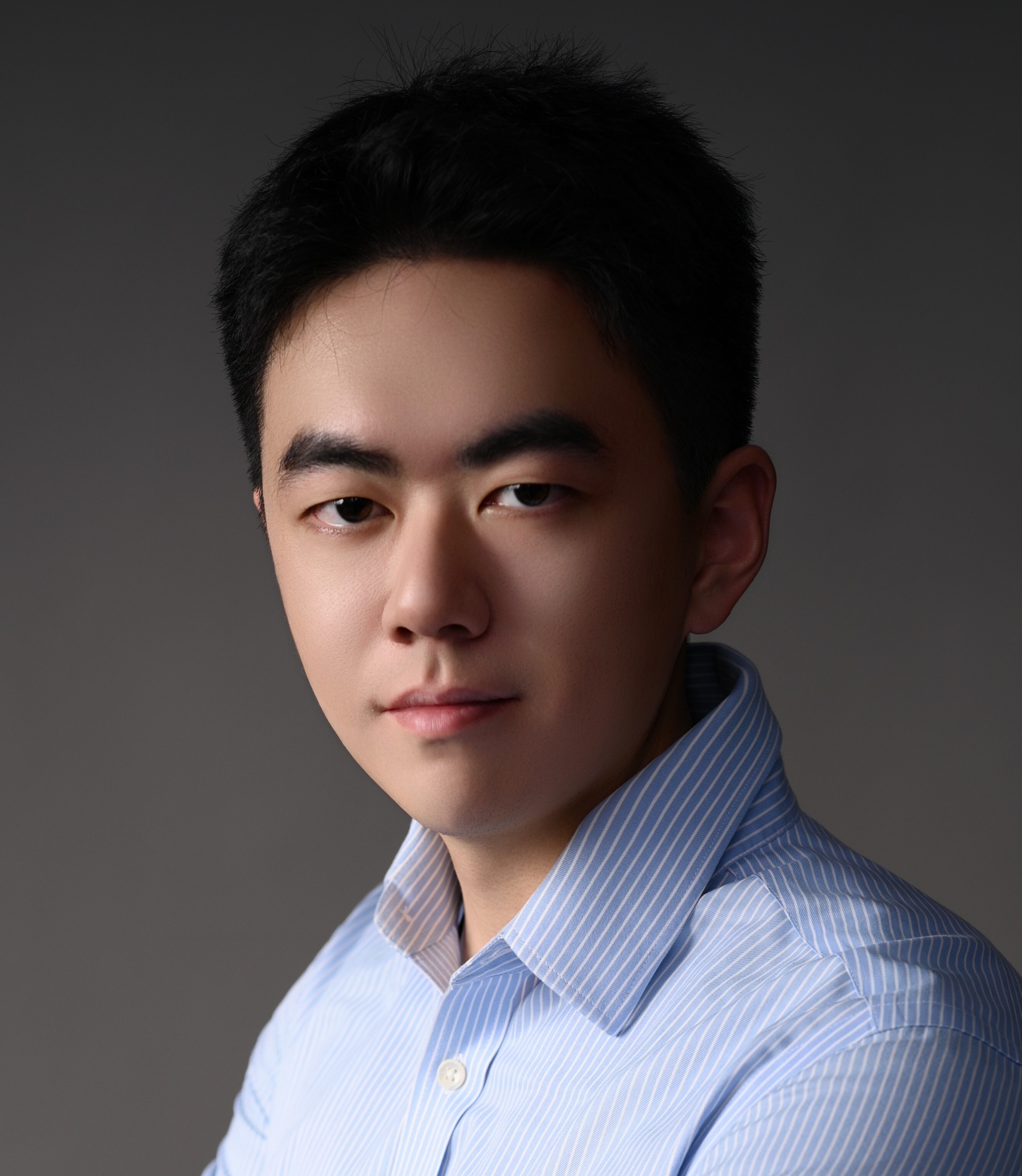}}]{Sangli Teng} is a postdoctoral researcher in Mechanical Engineering at the University of California, Berkeley. He received his Ph.D. in Robotics from the University of Michigan. His research leverages geometric principles to advance control, optimization, and learning for autonomous systems. He is a recipient of the Best Paper Award Finalist at the Robotics: Science and Systems (RSS) Conference 2023 and the Outstanding Reviewer Award at RSS 2025.
\end{IEEEbiography}

\vspace{-0.3 in}
\begin{IEEEbiography}
[{\includegraphics[width=1in,height=1.15in,clip]{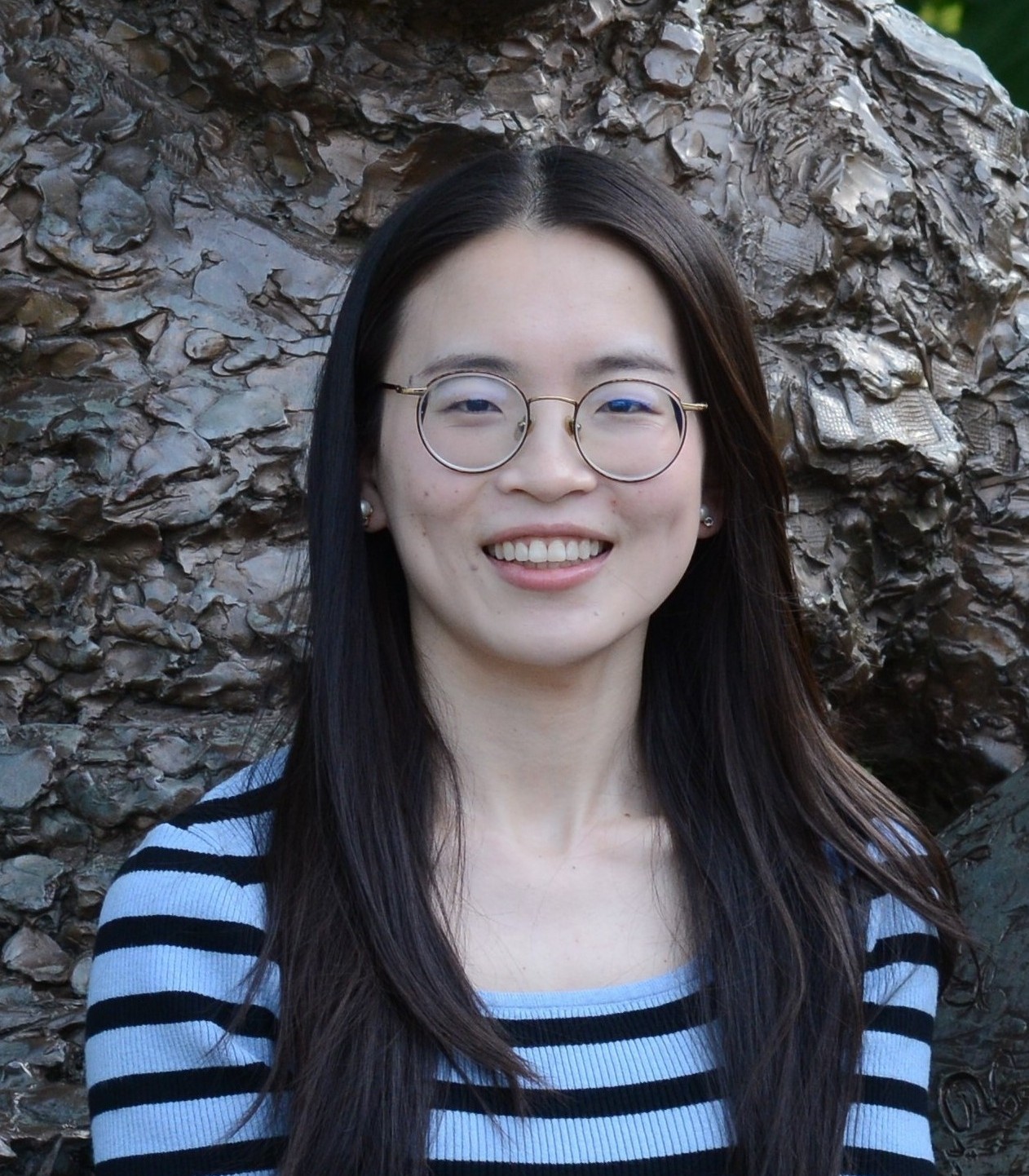}}]{Wenjing Li} received her B.S. degree in Mechanical Engineering from Huazhong University of Science and Technology in 2018 and her Ph.D. degree in Mechanical Engineering from the Georgia Institute of Technology, Atlanta, GA, USA in 2025. She is currently a Lillian Gilbreth Postdoctoral Fellow at Purdue University. Her research interests include mechatronics, actuator design and robotics.
\end{IEEEbiography}

\vspace{-0.3 in}

\begin{IEEEbiography}
[{\includegraphics[width=1in,height=1.15in,clip]{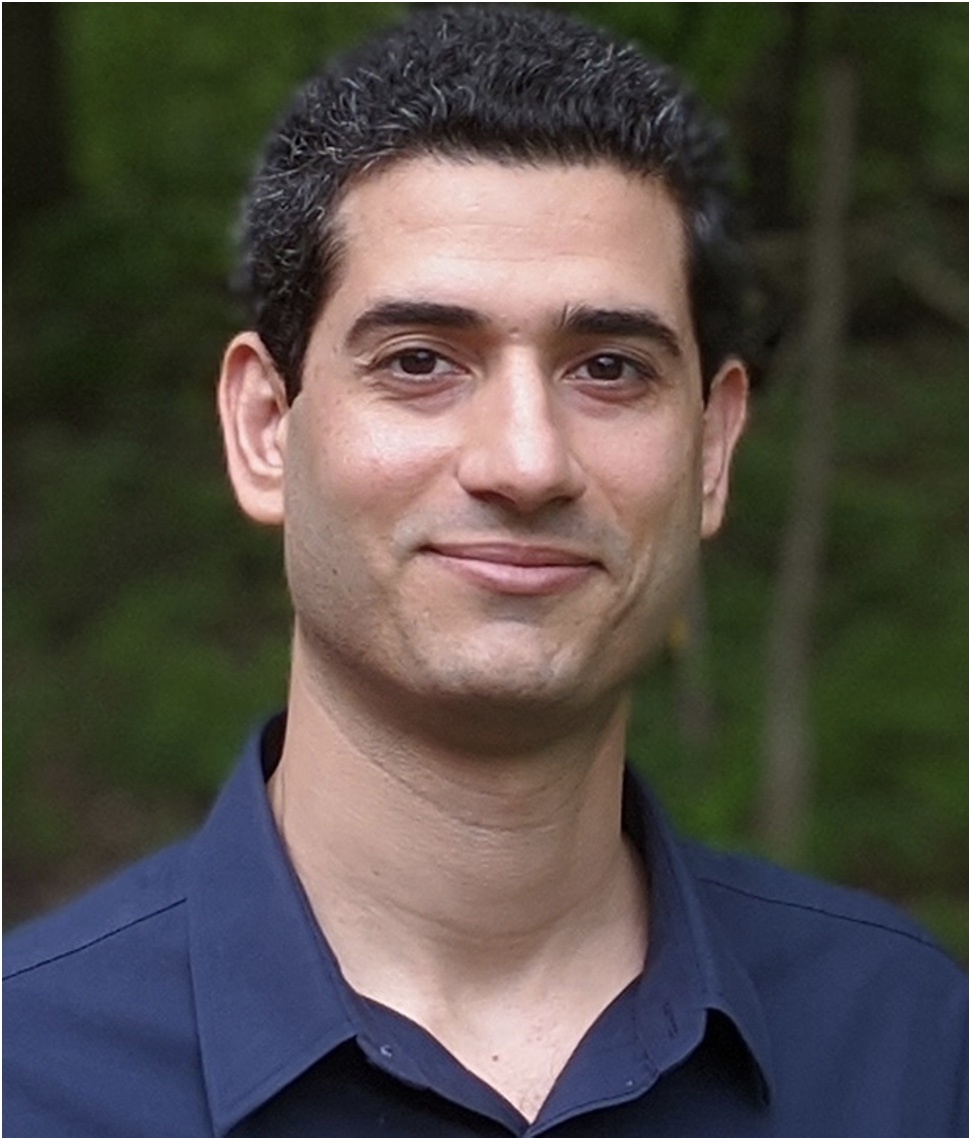}}]{Maani Ghaffari}
received the
Ph.D. degree from the Centre for Autonomous
Systems, University of Technology Sydney,
NSW, Australia, in 2017. He is currently an
Assistant Professor with the Department of Naval
Architecture and Marine Engineering and the
Department of Robotics, University of Michigan,
Ann Arbor, MI, USA. His research interests include the
theory and applications of robotics and autonomous systems. He was a recipient of the 2021 Amazon Research Awards.
\end{IEEEbiography}

\vspace{-0.3 in}

\begin{IEEEbiography}[{\includegraphics[width=1in,height=1.25in,clip,keepaspectratio]{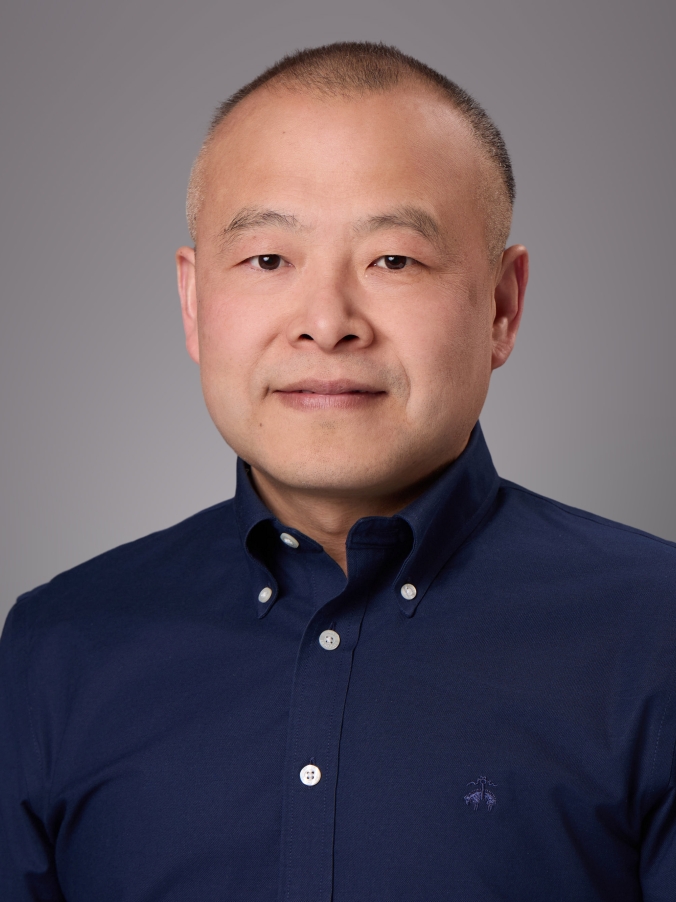}}]{Jingang Yi} received the Ph.D. degree in mechanical engineering from the University of California, Berkeley, CA, USA, in 2002. He is currently a Professor of mechanical engineering at Rutgers University. His research interests include autonomous robotic systems, dynamic systems and control, and mechatronics.
	
Dr. Yi is also a Fellow of the American Society of Mechanical Engineers (ASME). He was a recipient of the 2010 US NSF CAREER Award. He currently serves as a Senior Editor of the {\sc IEEE Transactions on Automation Science and Engineering} and served as a Senior Editor of the {\sc IEEE Robotics and Automation Letters} and a Technical Editor for {\sc IEEE/ASME Transactions on Mechatronics}.
\end{IEEEbiography}

\vspace{-0.3 in}

\begin{IEEEbiography}
[{\includegraphics[width=1in,height=1.15in,clip]{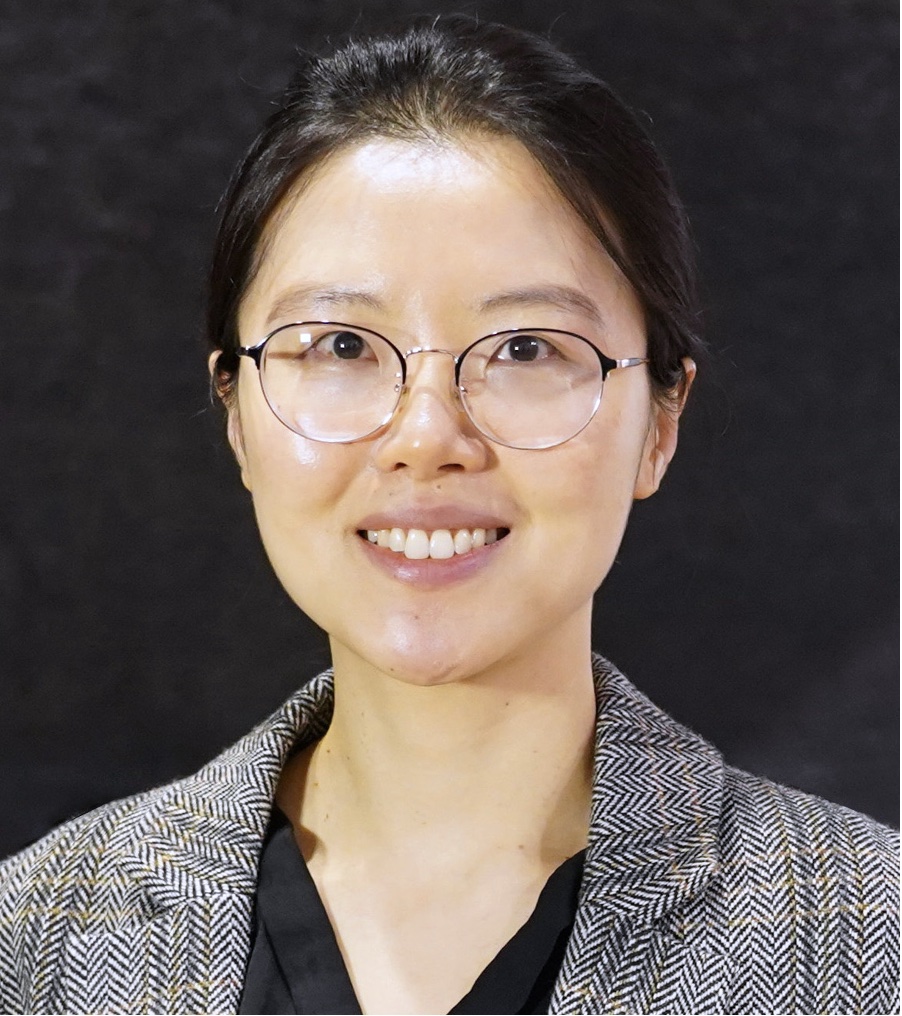}}]{Yan Gu}
received the B.S. degree in Mechanical Engineering from Zhejiang University in 2011 and the Ph.D. degree in Mechanical Engineering from Purdue University in 2017.
She joined the faculty of the School of Mechanical Engineering at Purdue University in 2022.
Her research interests include control theory, robot learning, humanoid robotics, and athletic intelligence.
She was the recipient of the NSF CAREER Award in 2021 and ONR YIP Award in 2023.
\end{IEEEbiography}

\end{document}

%% file: table/model_table.tex
\begin{table*}
\centering
\begin{threeparttable}

\caption{Representative robot models for legged locomotion in non-inertial environments.}
\label{tab:modeling}

\begin{tabularx}{\textwidth}{%
    >{\centering\arraybackslash}m{2cm}   % Column 1 width
    % >{\centering\arraybackslash}m{2cm}   % Column 2 width
    >{\centering\arraybackslash}m{3cm}   % Column 5 width
    >{\centering\arraybackslash}m{3cm}   % Column 5 width
    >{\centering\arraybackslash}m{6cm}   % Column 3 auto-expands
    >{\centering\arraybackslash}m{2cm}   % Column 4 width
}
\toprule
Robot model type &
Platform inertia (relative) and motion  &
Robot motion &
Key idea of robot dynamics modeling &
Robot model properties \\
% Environment motion type \\

\midrule

Reduced-order model & 
Lightweight; ball rolling~\cite{zheng_ball_2011, nagarajan_balancing_2014, kimura_extended_2021}, horizontal motion~\cite{konishi_zmp_2023}  & 
Bipedal standing~\cite{nagarajan_balancing_2014, kimura_extended_2021, konishi_zmp_2023}, bipedal walking~\cite{zheng_ball_2011} & 
Couple the simplified robot model with the dynamics model of the non-inertial environment through non-slip contact conditions & 
Linear, time-invariant 
\\

% Reduced-order model
& 
Heavy; 
horizontal motion \cite{chen_contingency_2024, gao_time-varying_2024},
vertical motion \cite{iqbal_analytical_2023, iqbal_asymptotic_2023},
horizontal and vertical motion \cite{stewart_adaptive_nodate} & 
Quadruped walking \cite{chen_contingency_2024, iqbal_analytical_2023, iqbal_asymptotic_2023}, 
bipedal walking \cite{gao_time-varying_2024, stewart_adaptive_nodate}
&
Model the non-inertial platform motion as time-varying parameters and/or disturbances & 
Linear, time-varying 
\\

\midrule

Full-order model & 
Lightweight platform: 6-D dynamic motion~\cite{bouyarmane_quadratic_2019}, Dynamic ball rolling~\cite{yang_dynamic_2020} &
Bipedal standing \cite{bouyarmane_quadratic_2019}, quadruped walking \cite{yang_dynamic_2020} & 
Couple the full-order robot model with the dynamics model of the non-inertial environment via holonomic contact constraints & 
Nonlinear, time-invariant \\

% Full-order model  
& 
Heavy; 6-D motion \cite{iqbal_provably_2020} &
Quadruped walking \cite{iqbal_provably_2020} &
Model the non-inertial effects through time-varying holonomic contact constraints & 
Nonlinear, time-varying \\

\midrule

Physics simulator 
&
Lightweight; ball rolling~\cite{ma_dreureka_2024} &
Quadruped walking~\cite{ma_dreureka_2024} &
Model the non-inertial environment as a passive ball under the robot&
Numerical simulation\\

&
Heavy; 
horizontal motion \cite{gao_time-varying_2024},
pitching motion \cite{iqbal_analytical_2023}, 6-D motion \cite{yoon_learning-based_2024}&
Quadruped walking \cite{iqbal_analytical_2023}, quadruped balancing \cite{yoon_learning-based_2024}, bipedal walking \cite{gao_time-varying_2024}
&
Model the non-inertial environment as an actuated plate under the robot
&
Numerical simulation\\

\bottomrule
\end{tabularx}

\end{threeparttable}
\end{table*}

%% file: table/control_table.tex
\begin{table}
\centering
\begin{threeparttable}

\caption{Representative works of control for legged locomotion in non-inertial environment}
\label{tab:control}

\begin{tabularx}{\columnwidth}{%
    >{\centering\arraybackslash}m{1.5cm}
    >{\centering\arraybackslash}m{2.2cm}
    >{\arraybackslash}X
}
\toprule
Control methodology &
Control objective &
Platform motion and knowledge\\
\midrule

Classical control &
Stable standing &
Horizontal; measured horizontal acceleration \cite{konishi_zmp_2023}\\

&
Stable walking&
Vertical; measured contact-point motion \cite{iqbal_provably_2020}\\

&
&
Horizontal; measured platform motion \cite{gao_time-varying_2024}\\[6pt]

&
&
Horizontal and vertical; known bound on motion frequency  \cite{stewart_adaptive_nodate}\\

\midrule

Optimization-based control &
Stable walking &
Horizontal; known bound on horizontal acceleration \cite{chen_contingency_2024}\\

&
&
Horizontal, roll and pitch; unknown \cite{gu_robust-locomotion-by-logic_2025}\\

&
&
Vertical; estimated vertical acceleration \cite{iqbal_ht-lip_nodate}\\

&
Ball position tracking&
Ball rolling; measured ball position \cite{yang_dynamic_2020}\\
\midrule

Reinforcement learning &
Velocity tracking &
Skateboarding; unknown \cite{liu_discrete-time_2025}\\

&
Stable standing  &
6-D motion; estimated platform motion \cite{yoon_learning-based_2024}\\

&
CoM position tracking&
Vertical; unknown \cite{zinage2026contractionppo}\\

&
Stable walking&
Horizontal; unknown \cite{da2021learning}\\

\bottomrule
\end{tabularx}

\end{threeparttable}
\end{table}